\documentclass[nohyperref]{article}

\usepackage{microtype}
\usepackage{graphicx}
\usepackage{subfigure}
\usepackage{xspace}
\usepackage{booktabs} % for professional tables
\usepackage{multirow}

\usepackage{hyperref}

\usepackage[accepted]{icml2022}

\usepackage{amsmath}
\usepackage{amssymb}
\usepackage{mathtools}
\usepackage{amsthm}

\setcitestyle{square,numbers,citesep={,}}

\usepackage[capitalize,noabbrev]{cleveref}

\theoremstyle{plain}

\theoremstyle{definition}

\theoremstyle{remark}

\usepackage[textsize=tiny]{todonotes}

\icmltitlerunning{Unsupervised Visual Relations Discovery with Graph-level Analogy}

\begin{document}

\twocolumn[
\icmltitle{ViRel: Unsupervised Visual Relations \\Discovery with Graph-level
Analogy}

\icmlsetsymbol{equal}{*}

\begin{icmlauthorlist}
\icmlauthor{Daniel Zeng}{equal,yyy}
\icmlauthor{Tailin Wu}{equal,yyy}
\icmlauthor{Jure Leskovec}{yyy}
\end{icmlauthorlist}

\icmlaffiliation{yyy}{Department of Computer Science, Stanford University, USA}

\icmlcorrespondingauthor{Daniel Zeng}{danielzeng@cs.stanford.edu}
\icmlcorrespondingauthor{Tailin Wu}{tailin@cs.stanford.edu}

\icmlkeywords{Machine Learning, ICML}

\vskip 0.3in
]

\printAffiliationsAndNotice{\icmlEqualContribution} % otherwise use the standard text.

\newcommand{\proj}{ViRel\xspace}
\begin{abstract}

Visual relations form the basis of understanding our compositional world, as relationships between visual objects capture key information in a scene. It is then advantageous to learn relations automatically from the data, as learning with predefined labels cannot capture all possible relations. However, current relation learning methods typically require supervision, and are not designed to generalize to scenes with more complicated relational structures than those seen during training. Here, we introduce \proj, a method for unsupervised discovery and learning of \textbf{Vi}sual \textbf{Rel}ations with graph-level analogy. In a setting where scenes within a task share the same underlying relational subgraph structure, our learning method of contrasting isomorphic and non-isomorphic graphs discovers the relations across tasks in an unsupervised manner. Once the relations are learned, \proj can then retrieve the shared relational graph structure for each task by parsing the predicted relational structure. Using a dataset based on grid-world and the Abstract Reasoning Corpus, we show that our method achieves above 95\% accuracy in relation classification, discovers the relation graph structure for most tasks, and further generalizes to unseen tasks with more complicated relational structures. Project website and code can be found at \url{http://snap.stanford.edu/virel/}.

\end{abstract}

\section{Introduction}

Our world is naturally compositional: where concepts, whether abstract or physical, are hierarchically composed of constituent concepts and their relations. In parallel, human intelligence has evolved to understand and reason with compositional structure. In the visual domain, this ability has endowed humans to quickly understand visual scenes and generalize to previously unseen, complex scenes. A key factor in compositionality is understanding visual relations, for example, that one object has the same shape as another.

\looseness=-1 A number of works have explored visual relations, for scene graph generation or image reconstruction \cite{DBLP:journals/corr/abs-1904-00560, DBLP:journals/corr/abs-1804-01622, DBLP:journals/corr/abs-1905-01608}, localization and grounding of subject-predicate-object triplets \cite{DBLP:journals/corr/LiOW17}, dynamics prediction \cite{DBLP:journals/corr/BattagliaPLRK16, https://doi.org/10.48550/arxiv.1911.12247}, visual relation detection \cite{DBLP:journals/corr/LuKBL16}, few-shot classification \cite{DBLP:journals/corr/abs-1711-06025}, and compound relation classification \cite{DBLP:journals/corr/abs-1905-10307}. These works have focused on applying visual relations to downstream higher-level tasks with known relation labels, or learning relations in a supervised setting.

In this work, we tackle a novel problem in visual relations, which is learning and discovering relations in the unsupervised setting, where relation types and labels are not known \emph{a priori}. In the unsupervised setting, the relevant relations are learned from the data, removing the dependence on predefined relation types. This unlocks greater generalization capacity for learning, reasoning and compositionality. In contrast, when relations are learned with predefined labels in a supervised context, this limits us to settings which depend on those seen relations.

Our key insight is that the emergence of relations comes from graph-level analogy, where scenes within a task share a common relational subgraph consisting of concepts as nodes and relations as edges. We introduce a graph-isomorphism based objective to learn a distinct graph embedding (computed from relation embeddings with a GNN) for each task, which then encourages distinct clusters of relation embeddings to form. Once our method has learned the relations, our method retrieves the shared relational graph structure for each task by parsing the predicted relational structures.

While the types and appearances of concepts within a task may vary, \proj is able to infer the global relation types, achieve above 95\% accuracy in relation classification in all our 6 dataset configurations, and retrieve the common relational graph structure in seen tasks and further generalize to unseen, more complex tasks.

\section{Task and Dataset}

\begin{figure}[ht]
\begin{center}
\centerline{\includegraphics[width=\columnwidth]{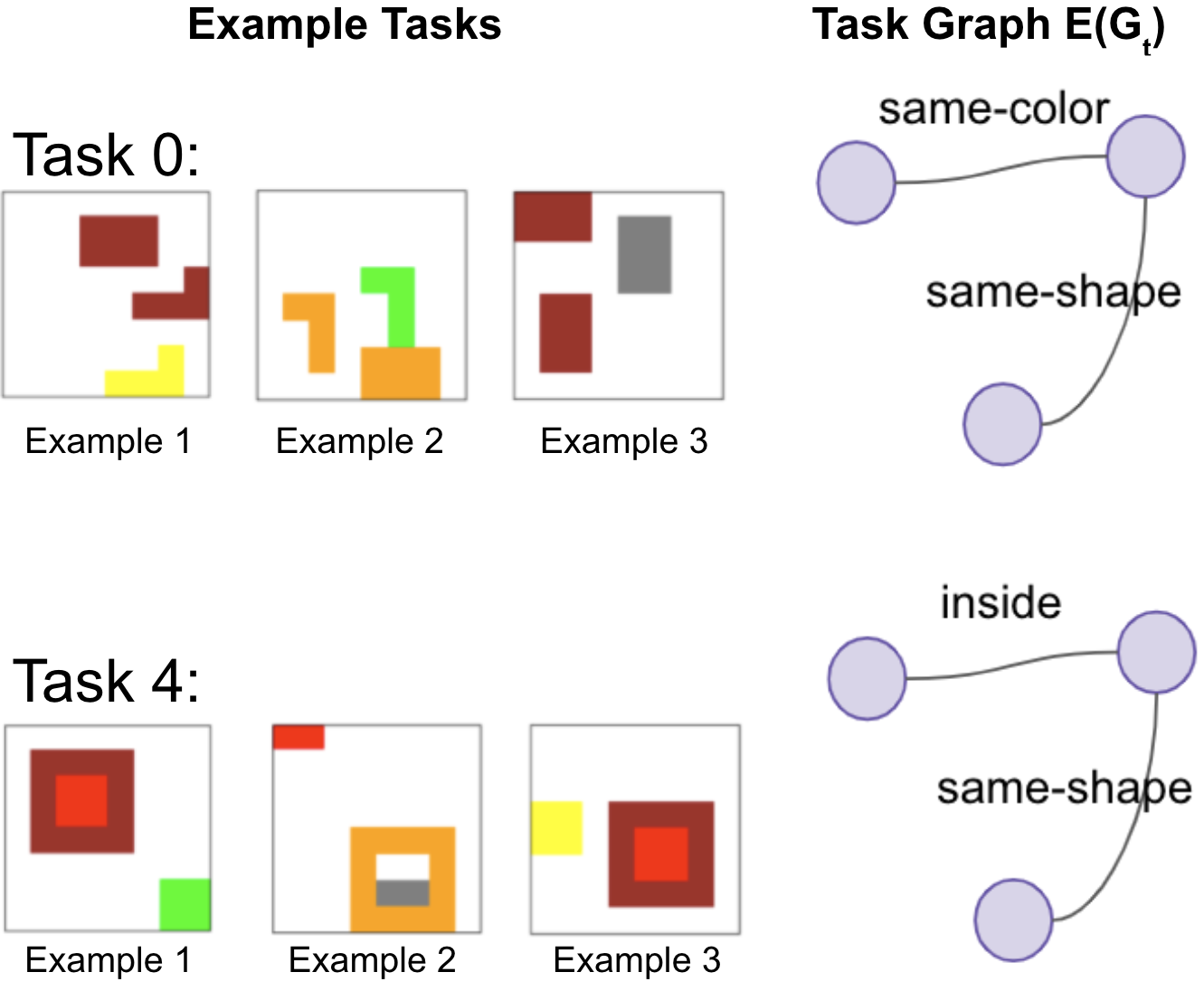}}
\end{center}
\vskip -0.3in
\caption{Two example BabyARC tasks and respective graphs. Each task $t$ contains images with a shared relation subgraph $E(G_t)$. The input are image examples under ``Example Tasks''. Both the relation types and the graph of each task are unknown. The goal is to infer the relation graph $E(g_i)$, the corresponding relations, and the shared relation graph $E(G_t)$ for each task $t$. More example inputs are shown in Fig. \ref{figure:bbarc_task_more}.}
\label{figure:sample_task}
\vskip -0.2in
\end{figure}

Our setup is inspired by the Abstraction and Reasoning Corpus (ARC) dataset \cite{DBLP:journals/corr/abs-1911-01547}, where the tasks provided in the dataset aim to serve as a benchmark for human-level intelligence and reasoning. One of the inductive priors in many ARC tasks is that the training examples are graph-isomorphic, where there exists a common relational structure between the examples. A motivating ARC task example is shown in Fig. \ref{figure:arc_task} and more in Fig. \ref{figure:arc_79} and \ref{figure:arc_76}. This follows our motivation of applying graph-level analogy, for the shared relational structure between tasks and to discover the relations between objects. 

However, due to the highly challenging and low data nature of ARC, we evaluate \proj on BabyARC \cite{wu2022zeroc}, a dataset generator which captures the graph isomorphic essence of ARC tasks. Moreover, BabyARC provides the underlying metadata of each generated example, allowing evaluation of our method, e.g., the accuracy of the predicted relations.

Datasets for visual reasoning such as CLEVR \cite{DBLP:journals/corr/JohnsonHMFZG16}, SHAPES \cite{DBLP:journals/corr/AndreasRDK15}, or Visual Question Answering \cite{DBLP:journals/corr/AntolALMBZP15,  DBLP:journals/corr/abs-1902-09506, DBLP:journals/corr/KrishnaZGJHKCKL16, https://doi.org/10.48550/arxiv.2205.00363} exist, but their setup differs greatly such as due to lack of configuration to specify shared graph structure, use of language query, their own additional task-specific complexities, or emphasis on different learning objectives.

\subsection{BabyARC Setup}
In the following, we introduce definitions for BabyARC:\\
\textbf{Definition 1.} Observation: represents a single image, which contains a collection of $n$ concepts, where $n$ is known.\\
\textbf{Definition 2.} Concept: represents an \textit{object} in the given observation. Concepts may be rectangles, lines, etc.\\
\textbf{Definition 3.} Relation: represents the relationship between two objects, specifically a visual relation in the BabyARC setting. For example, if two objects share the same color, the relation between these two objects would be referred to as ``same-color''. Similarly, ``same-shape'' represents that two objects have the same shape, and ``inside'' represents that one object is inside another object.  If two objects do not have any relations, the relation label is ``none''. 

Our task definition on the BabyARC is the following:\\
We are given a collection of observations (images) $x_i, i=1,2,\dots, N$, where each observation belongs to some known task $t \in T$. Each task $t$ has an unknown unique task graph $G_t = (V, E)$ with the nodes $V = \{$objects\} and $E = \{$relations between objects\}, and $e_{k, l} \in E$ represents the relation type between the $k^\text{th}$ and $l^\text{th}$ object. All of its corresponding observations share this common relational subgraph $E(G_t)$.

In addition, \emph{only the observations} of each task are provided to the model, \emph{without} knowledge of the objects, the global relation types, or the underlying graph of each observation. The goal is to infer the global relation types, the graph $E(g_i)$ of each observation $x_i$, and the relational graph structure $E(G_t)$ belonging to each task $t$. To accurately infer the underlying graph $E(g_i)$, this requires the model to identify the correct relation type $e_{k,l}$ between every object pair.

Two example BabyARC tasks are shown in Fig. \ref{figure:sample_task}. These examples show that the graph of each observation $E(g_i)$ is isomorphic to the task graph $E(G_t)$. However, it may be that only a subgraph of $E(g_i)$ is isomorphic to task graph $E(G_t)$. We define objects which are part of the relational subgraph as ``core'' objects, and ones which are not as ``distractor'' objects. Distractor objects can be viewed as random objects added to the observation and irrelevant to task graph $E(G_t)$. The distinction between distractor and core objects is not given to the model. Examples of tasks with distractor objects are shown in Fig. \ref{figure:sample_distractor}.

\section{Method}
\subsection{Concept-Relation Graph Neural Network (CR-GNN) Architecture}

\begin{figure*}[ht]
\vskip -0.1in
\begin{center}
\centerline{\includegraphics[trim={0 0.0cm 0 0.0cm},clip,width=1.5\columnwidth]{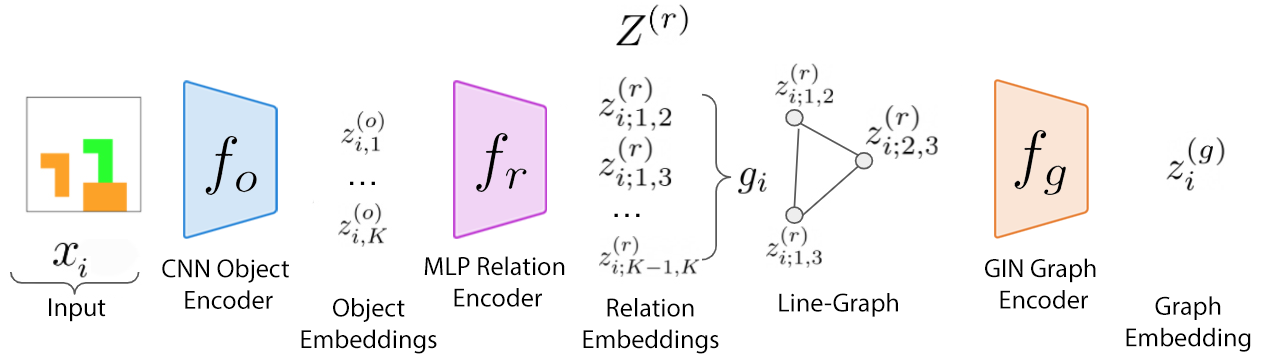}}
\caption{CR-GNN architecture, consisting of a CNN object encoder $f_o$, a MLP relation encoder $f_r$, and a GIN $f_g$.}
\label{figure:crgnn}
\end{center}
\vskip -0.3in
\end{figure*}

We propose the Concept-Relation Graph Neural Network (CR-GNN) architecture, which takes advantage of the subgraph isomorphic properties of each task. Our CR-GNN architecture consists of a CNN object encoder $f_o$ to encode each object, and outputs object embedding $z_{i,k}^{(o)}$ for the $k^\text{th}$ object from image $x_i$, a MLP relation encoder $f_r$ to encode the relation $z_{i;k,l}^{(r)}$ between the $k^\text{th}$ and $l^\text{th}$ object.

Encoding the input image through the object and relation encoders, the image $x_i$ is represented as a latent graph $g_i$, with $z_{i;k,l}^{(r)}$ being its node features (note that here we use the line-graph representation, where the node features are the relation embedding), and two nodes are connected if two relations shares the same object. After, a 2 layer $f_g$  graph isomorphism network (GIN) \cite{xu2018powerful} is applied to encode the learned line-graph. An illustration of the CR-GNN architecture is shown in Fig. \ref{figure:crgnn}.

The model design choices are guided by the goal of the task objective. The CNN object encoder embeds each visual object into an embedding vector, and the MLP relation encoder maps each pairwise concatenation of objects into a relation embedding vector. We output an explicit representation for the relation vector encoding as this forces the GNN encoder to depend only on the relational properties of the input, rather than other properties of the objects themselves. We apply a GIN encoder on top of the line-graph constructed with the relation latent vectors to take advantage that all observations of the same task are subgraph-isomorphic. 

\subsection{Loss function objective}

We introduce and explore two main loss objectives to learn the CR-GNN model. The first objective is a contrastive learning based objective which minimizes the distance of the graph representation of each example of the same task (intra-task), while maximizing the distance with respect to examples not in the same task (inter-task). The mathematical formulation of the contrastive objective is:

\vskip -0.3in
\label{eq:gin_objective}
\begin{multline}
    \mathcal{L}_\text{contrastive} = \sum_{i, j \in \text{same-task}} ||f_g(g_i) - f_g(g_j)||_2 \\
    + \sum_{k, m \in \text{diff-task}} \max(0, \eta - ||f_g(g_k) - f_g(g_m)||_2
\end{multline}
where the first summation defines the intra-task loss, and the second summation defines the inter-task loss. Given that each task shares a common subgraph, its graph representation $f_g(g_i)$ should be similar within the task (intra-task loss), and should be different between different tasks (inter-task loss). $\eta$ here is a margin hyperparameter.

The second objective is a classification (cross-entropy) based objective. Each example is classified as one of task graphs, using the true task label as ground truth. The mathematical formulation of the classification objective is:

\vskip -0.3in
\begin{align}
\label{eq:classify_objective}
 \mathcal{L}_\text{classify} = \sum_{\forall i \in n} \mathcal{L}_{CE}( \text{Linear}(f_g(g_i)), y)
\end{align}
\vskip -0.2in
where $\mathcal{L}_{CE}$ is the standard cross-entropy loss, between the true task ID $y$ against the predicted task ID. The predictions logits are obtained via a linear layer following the graph representation learned from the GIN $f_g$. This linear layer is trained along-side the other model components.

An additional loss term, which is the information bottleneck loss, allows constraining the information between the observation $X$ and the relation embedding $Z^{(r)}$, which forces the relation latent dimension to use more clustered embedding. This can be seen as a regularization on $Z^{(r)}$ to capture only the most relevant information about $X$, which is the relation types in our case. The mathematical formulation of the information bottleneck (IB) is:

\vskip -0.7cm
\begin{align}
\label{eq:ib_term}
 \mathcal{I}(X;Z^{(r)})
\end{align}
\vskip -0.4cm

where $\mathcal{I}$ is the mutual information function (Eq. \ref{eq:mi_term}), $Z^{(r)}$ is the random variable concatenating  the relation embedding $z_{i;k,l}^{(r)}$ for an observation $x_i \in X$. This implementation follows from \citet{DBLP:journals/corr/AlemiFD016} which uses a variational upper bound  \cite{https://doi.org/10.48550/arxiv.1312.6114} for tractable computation.

\section{Experiments}
\subsection{Setup}

In the following experiments, the global relation types for BabyARC dataset are defined to be ``none'', ``same-shape'', ``same-color'', and ``inside''. Visual examples of these relations are shown in Fig. \ref{figure:relations}.  The object shapes are a rectangle (``rect-solid''), hollow rectangle (``rect''), ``Lshape'', and line ``line'', and see Fig. \ref{figure:sample_concepts} for some examples. 

Our dataset specifications also allow us to generate datasets of varying numbers of core objects and distractor objects. We investigate two categories of datasets: tasks containing 2-3 core objects, and tasks containing 2-4 core objects. We also vary the number of distractor objects, with three configurations: no distractor objects, 1 distractor object, and 0-2 distractor objects. The relation specification for each task are described in Appendix \ref{appendix:ds_task}.

Examples of two tasks of the (2-3 core, 0 distractor) generated dataset is shown in Fig. \ref{figure:sample_task}. In the first image of Task 0, the brown rectangle has the relation “same-color” with the brown “Lshape”, and the brown “Lshape” has the relation “same-shape” as the yellow “Lshape”. 

We train our CR-GNN model with the two loss objectives as previously defined, and on datasets with 2-3 core objects or 2-4 core objects, and varying number of distractor objects depending on the configuration. We also observed the effect of adding the IB loss. Our training and model hyperparameters are described in Appendix \ref{appendix:arch}.

\subsection{Results}

\begin{table}[t]
\caption{Relation classification accuracy for 2-3 core objects}
\label{table:3obj}
\begin{center}
\begin{small}
\begin{sc}
\begin{tabular}{|l|c|c|c|}
\hline
Method  & \multicolumn{3}{|l|}{\# Distractors}  \\ \hline
  & 0 & 1   & 0-2  \\ \hline
Classify        & 0.923          & 0.926 & 0.946  \\ \hline
Classify + IB    & 0.919 & 0.918          &   0.901              \\ \hline
Contrastive     & \textbf{0.959}          & 0.961          & 0.954                 \\ \hline
Contrastive + IB & 0.952          & \textbf{0.963}                & \textbf{0.957}                 \\ \hline
Best            & \textbf{0.959} & \textbf{0.963} & \textbf{0.957}  \\ \hline
\end{tabular}
\end{sc}
\end{small}
\end{center}
\vskip -0.3in
\end{table}

The model is evaluated on two main aspects: the accuracy of the model with predicting the correct relation type between two objects, and inferring the relational graph structure belonging to each task. These evaluations are based off the dataset task goals we have defined in Section 2.

We evaluate the relation prediction accuracy as follows: we apply \textit{k}-means clustering to assign cluster labels to each of the learned relation embeddings $z_{i;k,l}^{(r)}$. We permute globally how each cluster label is assigned to the ground-truth relation label.
The maximum accuracy with respect to the ground truth label $e_{k,l}$ is then taken. We only compute the relation accuracy between objects which contain a relation label, as we do not know the underlying relation of object pairs without labels. Model accuracy evaluation is done on a validation dataset with the same parameters used to generate the training dataset, but with different random seed.

Table \ref{table:3obj} \& \ref{table:4obj} show the relation classification accuracy for our configurations. The model performance is similar between both objectives, with the contrastive objective performing slightly better. In varying the number of introduced distractor objects in both cases, there is not any accuracy degradation due to this introduction.

\begin{table}[t]
\caption{Relation classification accuracy for 2-4 core objects}
\label{table:4obj}
\begin{center}
\begin{small}
\begin{sc}
\begin{tabular}{|l|c|c|c|}
\hline
Method & \multicolumn{3}{|l|}{\# Distractors}  \\ \hline
 & 0 & 1   & 0-2  \\ \hline
Classify        & 0.956          & 0.955      & 0.965       \\ \hline
Classify + IB    & 0.960 & 0.962                & 0.959  \\ \hline
Contrastive     &   \textbf{0.965}             &   0.971             &   0.965              \\ \hline
Contrastive + IB & 0.960        &  \textbf{0.973 }               & \textbf{0.971}                \\ \hline
Best            & \textbf{0.965} & \textbf{0.973} & \textbf{0.971}  \\ \hline
\end{tabular}
\end{sc}
\end{small}
\end{center}
\vskip -0.3in
\end{table}

Comparing the 2-3 core objects dataset against the 2-4 core objects dataset, there is an overall slight improvement with training on the 2-4 core objects dataset. This is because the 2-4 core objects dataset includes a greater number of task examples (13 vs 6), which allows learning better relation representations in order for the model to distinguish between different graphs.

Our model is able to infer the global relation types, as shown in Fig. \ref{figure:tsne-rel}, a t-SNE visualization of the learned relation embeddings. The model clusters relation embeddings of the same relation label close to each other, even when not given the ground truth relation labels.

To infer the relational graph structure belonging to each task, \textit{k}-means evaluation method is used to predict the cluster labels of the learned relation. The graph for each observation is then constructed with these predicted labels. Since the number of objects in the observations of each task may be different due to distractors, we take the maximum common subgraph (MCS) of all the same-task constructed graphs, which allows identifying the shared relational graph belonging to each task. 

Our MCS retrievals for the 2-4 core object, 0 distractor dataset with the contrastive + IB objective are in Appendix \ref{appendix:task_mcs_preds}, with evaluation method details described. \proj is trained and evaluated on the same 2-4 core object, 0 distractor dataset. The top 3 most frequently obtained maximum common subgraphs for each task are shown, and the ground truth relational graph occurs in the top 3 retrievals for most tasks. This demonstrates that \proj is able to uncover the ground truth shared relational graph structure for most tasks.

Furthermore, we demonstrate the ability of our model to generalize to tasks unseen during training. We train \proj on the 2-3 core object, 1 distractor dataset with the contrastive + IB objective, and at inference time, evaluate MCS retrievals on the 2-4 core object, 0 distractor dataset. The relational structure of Tasks 6 to 12 are not seen during training, but obtains the correct MCS in the top 3 retrievals for most tasks. The full retrievals are shown in Appendix \ref{appendix:task_mcs_preds_unseen}.

In this work, we propose \proj for unsupervised learning and discovery of visual relations with graph-level analogy. We show that \proj is able to infer the global relation types, achieve above 95\% accuracy in relation classification, and retrieve the shared relational graph structure for most tasks.

One limitation of \proj is that it only learns the necessary relation representations needed to distinguish between the given tasks, and thus does not distinguish between relations without a label ``none``, and the ``same-color`` relation. In the MCS retrievals, extraneous predictions of ``same-color`` are found due to this reason. Future work would be understanding how to separate these two representations, with a promising approach being introducing more diverse tasks.

\section{Acknowledgement}
We thank Rok Sosič for discussions and for providing feedback on our manuscript.
We also gratefully acknowledge the support of
DARPA under Nos. HR00112190039 (TAMI), N660011924033 (MCS);
ARO under Nos. W911NF-16-1-0342 (MURI), W911NF-16-1-0171 (DURIP);
NSF under Nos. OAC-1835598 (CINES), OAC-1934578 (HDR), CCF-1918940 (Expeditions), 
NIH under No. 3U54HG010426-04S1 (HuBMAP),
Stanford Data Science Initiative, 
Wu Tsai Neurosciences Institute,
Amazon, Docomo, GSK, Hitachi, Intel, JPMorgan Chase, Juniper Networks, KDDI, NEC, and Toshiba.

The content is solely the responsibility of the authors and does not necessarily represent the official views of the funding entities.

\nocite{langley00}

\bibliography{example_paper}
\bibliographystyle{icml2022}

\newpage
\appendix
\onecolumn
\section{Appendix}

\begin{figure}[ht]
\vskip 0.2in
\begin{center}
\centerline{\includegraphics[trim={0 1.3cm 0 0},clip,width=0.5\columnwidth]{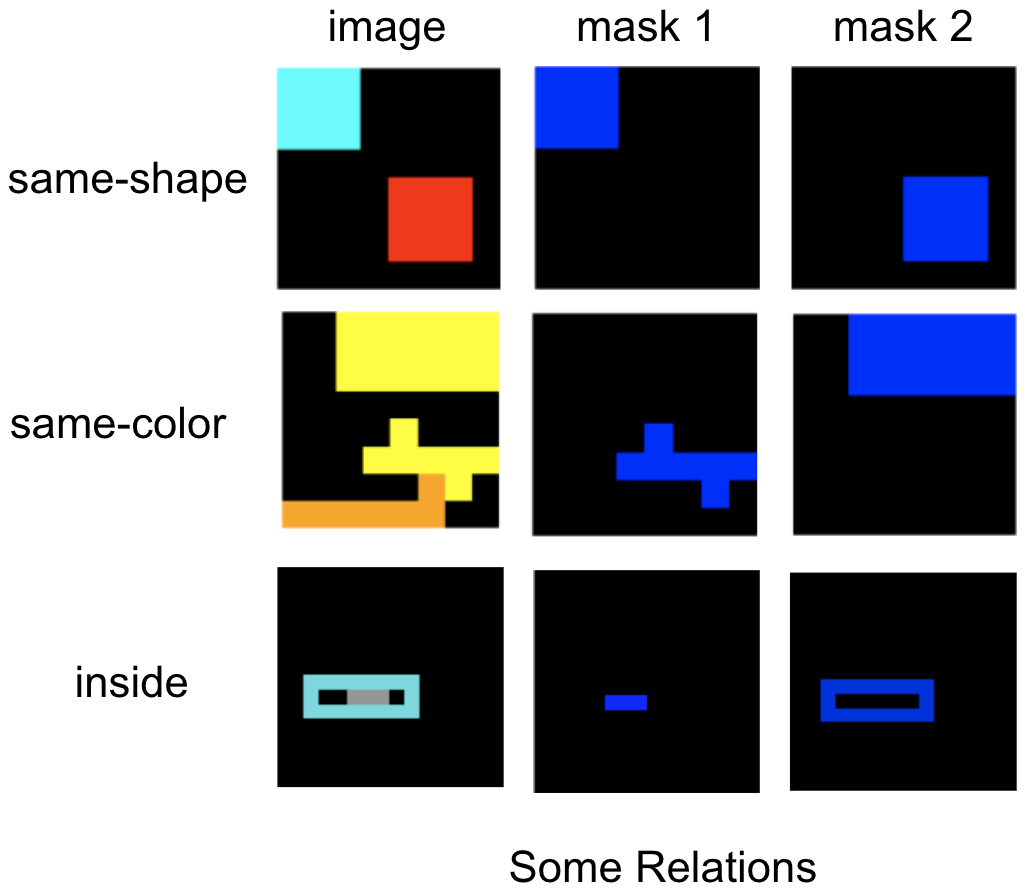}}
\caption{Example of each of the three global relation types of the BabyARC dataset.}
\label{figure:relations}
\end{center}
\vskip -0.2in
\end{figure}

\begin{figure}[ht]
\begin{center}
\centerline{\includegraphics[width=0.5\columnwidth]{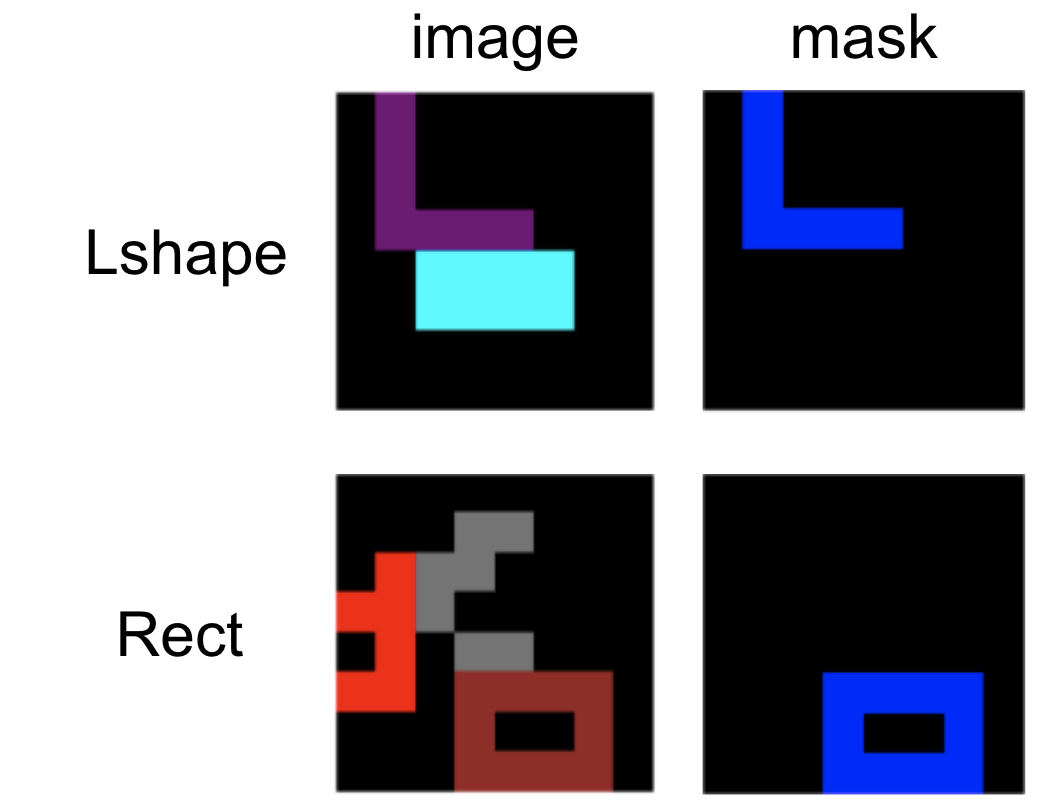}}
\caption{Concepts: Examples of ``Lshape`` and ``Rect``.}
\label{figure:sample_concepts}
\end{center}
\vskip -0.3in
\end{figure}

\begin{figure}[h]
\begin{center}
\centerline{\includegraphics[width=0.6\columnwidth]{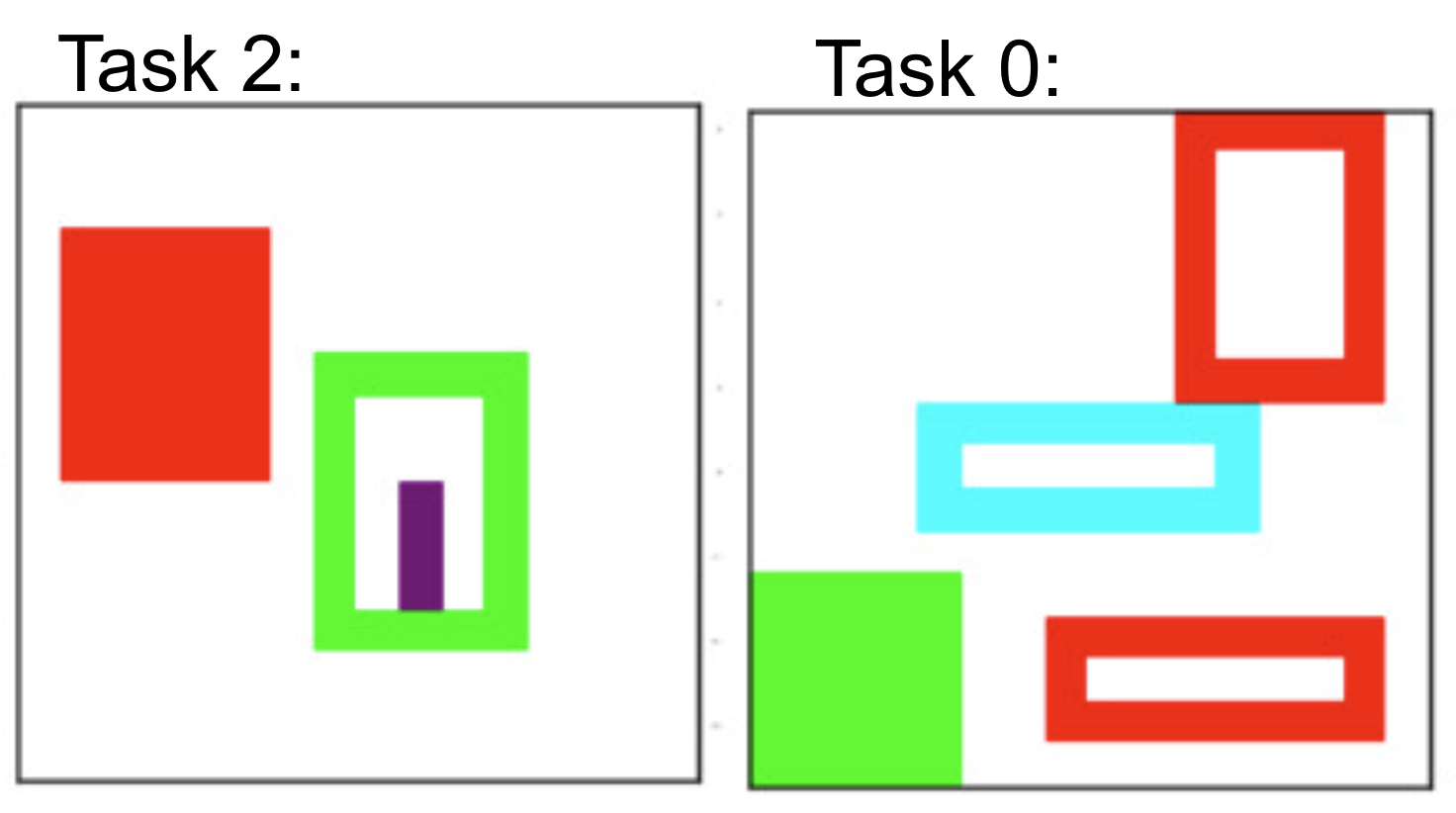}}
\caption{Example of Task 2 (2 core objects), with red rectangle distractor object, and Task 0 (3 core objects), with green distractor object. The core objects are part of the relational subgraph.}
\label{figure:sample_distractor}
\end{center}
\vskip -0.3in
\end{figure}

\begin{figure}[h]
\begin{center}
\centerline{\includegraphics[trim={0 10cm 0 0cm},clip,width=0.5\columnwidth]{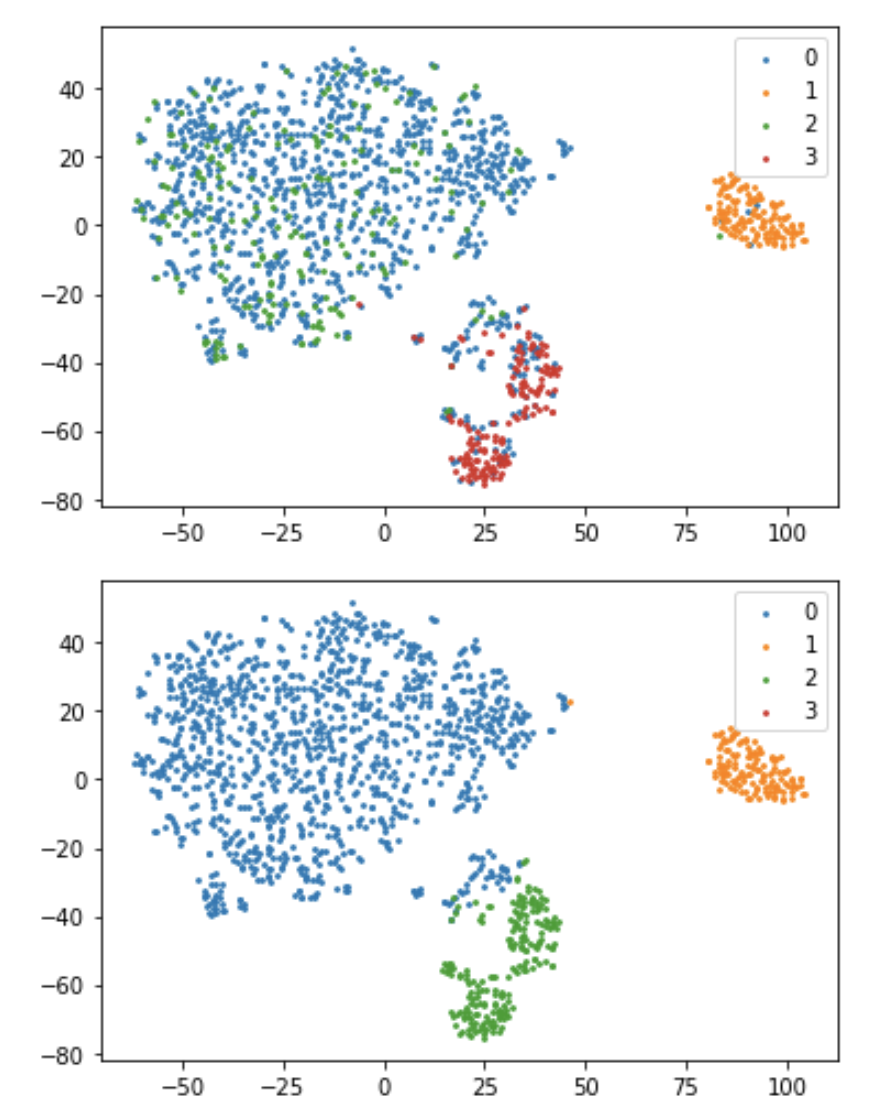}}
\caption{t-SNE visualization of learned relation embeddings, for 2-3 core, 0-2 distractor dataset with contrastive + IB objective. The colors represent ground truth labels, where label 0 represents no relationship label ``none'', 1 is ``inside'', 2 is ``same-color'', 3 is ``same-shape''.}
\label{figure:tsne-rel}
\end{center}
\end{figure}

\subsection{Mutual Information Function}
The mutual information for PDFs for continuous distributions is defined as the following:\\
\begin{align}
\label{eq:mi_term}
 \mathcal{I}(X,Y)=\int_y \int_x p(y,x)\log\frac{p(y,x)}{p(y)p(x)} dx\,dy\,
\end{align}

\subsection{Hyperparameters: Model Architecture and Training}
\label{appendix:arch}
The CNN object encoder is a 4 layer CNN, MLP relation encoder is a 3 layer MLP, and GIN graph encoder is 2 layer GNN with 3 layer MLP for each GNN layer. All the activation functions used are the LeakyRelu function. The model was optimized using the Adam \cite{DBLP:journals/corr/KingmaB14} optimizer with $1 \times 10^{-4}$ learning rate, 0.9 momentum, and no weight decay. The latent dimension for objects is 100, and the latent dimension for relations is 20. The input with is an image of $16 \times 16 \times 9 \times n$, where $16 \times 16$ is width by height, 9 is the number of total colors, and $n$ is the maximum number of possible objects in the dataset. The ground truth object masks are provided as input, as our work is focused on relation discovery, and not on object discovery. The margin hyperparameter $\eta$ for the contrastive loss is $20 \times \frac{2}{3}$, where 20 is the relation latent dimension. Each image is augmented with probability $0.9$, where the possible augmentations are random flipping, rotating, resizing, and color. The weighting for each respective losses are $1.0$ for contrastive, if used, $1.0$ for classify, if used, and $0.1$ for IB, if used. In training, each task $t$ contains around 200-300 image examples.

\subsection{Limitations}
One current limitation is that the model only learns the necessary relation representations needed to distinguish between the given tasks. We observed this similarly in our earlier experiments, where we evaluated our method in a setting with a limited number of tasks, and our method only learned two relation representations as it was sufficient to separate the tasks: ``inside`` and ``same-shape\textbackslash same-color``. 

Our method currently does not distinguish between relations without a label ``none``, and the ``same-color`` relation. This can be visualized in the t-SNE visualization in Fig. \ref{figure:tsne-rel}, where the ``none`` and ``same-color`` relation distributions overlap with each other.

\subsection{Tasks in each Dataset}
\label{appendix:ds_task}
\begin{figure}[H]
\begin{center}
\centerline{\includegraphics[trim={0 1400 0 0},clip,width=\columnwidth]{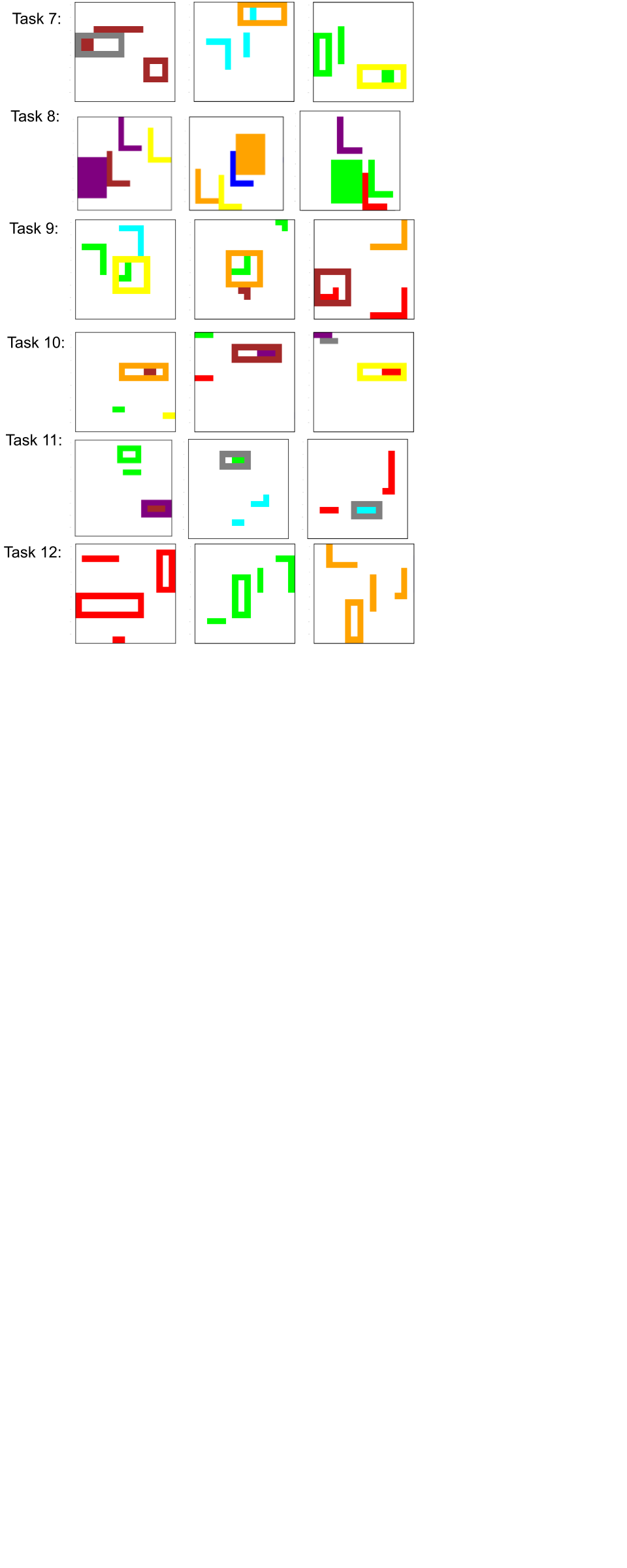}}
\caption{BabyARC: The images above depict 3 examples for each task. While only 3 are shown for each task, we use 200-300 examples per task during training.}
\label{figure:bbarc_task_more}
\end{center}
\vskip -0.3in
\end{figure}

The notation is the following: [(0, 1), 'same-color'] represents that the relation 'same-color' holds between the 0th and the 1st object. Only core objects are defined in the relations specification, not distractor objects. Visual examples for each task are shown in Fig. \ref{figure:bbarc_task_more}.

For datasets containing 2-3 core objects, the tasks in the dataset are the following:

\noindent \textbf{Task 0} (3 objects): [(0, 1), 'same-color'], [(1, 2), 'same-shape']\\
\textbf{Task 1} (2 objects): [(0, 1), 'same-color']\\
\textbf{Task 2} (2 objects): [(0, 1), 'inside']\\
\textbf{Task 3} (3 objects): [(0, 1), 'inside'], [(1, 2), 'same-color']\\
\textbf{Task 4} (3 objects): [(0, 1), 'inside'], [(1, 2), 'same-shape']\\
\textbf{Task 5} (3 objects): [(0, 1), 'same-color'], [(1, 2), 'same-color'] \\

For datasets containing 2-4 core objects, the tasks in the dataset are the following:

\noindent\textbf{Task 0} (2 objects): [(0, 1), 'inside']\\
\textbf{Task 1} (3 objects): [(0, 1), 'inside'], [(1, 2), 'same-color']\\
\textbf{Task 2} (3 objects): [(0, 1), 'inside'], [(1, 2), 'same-shape']\\
\textbf{Task 3} (2 objects): [(0, 1), 'same-color']\\
\textbf{Task 4} (3 objects): [(0, 1), 'same-color'], [(1, 2), 'same-color']\\
\textbf{Task 5} (3 objects): [(0, 1), 'same-color'], [(1, 2), 'same-shape']\\
\textbf{Task 6} (4 objects): [(0, 2), 'same-color'], [(1, 2), 'same-color'], [(2, 3), 'same-shape']\\
\textbf{Task 7} (4 objects): [(0, 1), 'inside'], [(1, 2), 'same-color'], [(2, 3), 'same-color']\\
\textbf{Task 8} (4 objects): [(0, 2), 'same-shape'], [(1, 2), 'same-color'], [(2, 3), 'same-shape']\\
\textbf{Task 9} (4 objects): [(0, 1), 'inside'], [(1, 2), 'same-color'], [(2, 3), 'same-shape']\\
\textbf{Task 10} (4 objects): [(0, 1), 'inside'], [(1, 2), 'same-shape'], [(2, 3), 'same-shape']\\
\textbf{Task 11} (4 objects): [(0, 1), 'inside'], [(1, 2), 'same-shape'], [(2, 3), 'same-color']\\
\textbf{Task 12} (4 objects): [(0, 2), 'same-color'], [(1, 2), 'same-color'], [(2, 3), 'same-color']\\

\subsection{Task Maximum Common Subgraph Retrieval, with all seen tasks}
\label{appendix:task_mcs_preds}

The following show the Maximum Common Subgraph Retrievals for \proj trained on 2-4 core object, 0 distractor dataset with the contrastive + IB objective. The evaluation dataset is the same, on 2-4 core object, 0 distractor dataset. Thus, relational structure of all the tasks shown is seen by the model during training.

To obtain the shared relational graph structure for each task, we would ideally take the maximum common subgraph of all the constructed graphs on each task. In practice, however, the relation predictions are not perfect and one incorrect relation prediction would then change the whole maximum common subgraph. To mitigate this, we compute maximum common subgraph of subgroups instead of all the observations at once, and then use a counting mechanism to identify the most frequently retrieved maximum common subgraph. The size of these subgroups is referred as the "group size". In the following result, the group size used is 5. The notation for interpreting the result is same as in Section \ref{appendix:ds_task}.

\textbf{Task 0:}\\
\textbf{Count:} 420, \textbf{MCS:} [(0, 1), 'inside']

\textbf{Task 1:}\\
\textbf{Count:} 287, \textbf{MCS:} [(0, 1), 'inside'], [(0, 2), 'same-color'], [(1, 2), 'same-color']\\
\textbf{Count:} 133, \textbf{MCS:} [(0, 1), 'inside'], [(1, 2), 'same-color']

\textbf{Task 2:}\\
\textbf{Count:} 337, \textbf{MCS:} [(0, 1), 'inside'], [(0, 2), 'same-color'], [(1, 2), 'same-shape']\\
\textbf{Count:} 83, \textbf{MCS:} [(0, 1), 'inside'], [(1, 2), 'same-shape']

\textbf{Task 3:}\\
\textbf{Count:} 420, \textbf{MCS:} [(0, 1), 'same-color']

\textbf{Task 4:}\\
\textbf{Count:} 270, \textbf{MCS:} [(0, 1), 'same-color'], [(0, 2), 'same-color'], [(1, 2), 'same-color']\\
\textbf{Count:} 150, \textbf{MCS:} [(0, 1), 'same-color'], [(0, 2), 'same-color']

\textbf{Task 5:}\\
\textbf{Count:} 305, \textbf{MCS:} [(0, 1), 'same-color'], [(0, 2), 'same-shape'], [(1, 2), 'same-color']\\
\textbf{Count:} 115, \textbf{MCS:} [(0, 1), 'same-color'], [(0, 2), 'same-shape']

\textbf{Task 6:}\\
\textbf{Count:} 159, \textbf{MCS:} [(0, 1), 'same-color'], [(0, 2), 'same-color'], [(0, 3), 'same-color'], [(1, 2), 'same-color'], [(1, 3), 'same-color'], [(2, 3), 'same-shape']\\
\textbf{Count:} 144, \textbf{MCS:} [(0, 1), 'same-color'], [(0, 2), 'same-color'], [(0, 3), 'same-color'], [(1, 2), 'same-color'], [(2, 3), 'same-shape']\\
\textbf{Count:} 78, \textbf{MCS:} [(0, 1), 'same-color'], [(0, 2), 'same-color'], [(1, 2), 'same-color'], [(2, 3), 'same-shape']

\textbf{Task 7:}\\
\textbf{Count:} 230, \textbf{MCS:} [(0, 1), 'inside'], [(0, 2), 'same-color'], [(0, 3), 'same-color'], [(1, 3), 'same-color'], [(2, 3), 'same-color']\\
\textbf{Count:} 101, \textbf{MCS:} [(0, 1), 'inside'], [(0, 2), 'same-color'], [(0, 3), 'same-color'], [(1, 2), 'same-color'], [(1, 3), 'same-color'], [(2, 3), 'same-color']\\
\textbf{Count:} 69, \textbf{MCS:} [(0, 1), 'inside'], [(0, 2), 'same-color'], [(0, 3), 'same-color'], [(1, 2), 'same-color']

\textbf{Task 8:}\\
\textbf{Count:} 242, \textbf{MCS:} [(0, 1), 'same-color'], [(0, 2), 'same-shape'], [(0, 3), 'same-shape'], [(1, 2), 'same-color'], [(1, 3), 'same-color'], [(2, 3), 'same-shape']\\
\textbf{Count:} 110, \textbf{MCS:} [(0, 1), 'same-color'], [(0, 2), 'same-shape'], [(0, 3), 'same-shape'], [(1, 2), 'same-color'], [(2, 3), 'same-shape']\\
\textbf{Count:} 39, \textbf{MCS:} [(0, 1), 'same-color'], [(0, 2), 'same-shape'], [(0, 3), 'same-shape'], [(2, 3), 'same-shape']

\textbf{Task 9:}\\
\textbf{Count:} 195, \textbf{MCS:} [(0, 1), 'inside'], [(0, 2), 'same-color'], [(0, 3), 'same-color'], [(1, 2), 'same-color'], [(1, 3), 'same-color'], [(2, 3), 'same-shape']\\
\textbf{Count:} 112, \textbf{MCS:} [(0, 1), 'inside'], [(0, 2), 'same-color'], [(0, 3), 'same-color'], [(1, 2), 'same-color'], [(2, 3), 'same-shape']\\
\textbf{Count:} 97, \textbf{MCS:} [(0, 1), 'inside'], [(0, 2), 'same-color'], [(0, 3), 'same-color'], [(2, 3), 'same-shape']

\textbf{Task 10:}\\
\textbf{Count:} 253, \textbf{MCS:} [(0, 1), 'inside'], [(0, 2), 'same-color'], [(0, 3), 'same-color'], [(1, 2), 'same-shape'], [(1, 3), 'same-shape'], [(2, 3), 'same-shape']\\
\textbf{Count:} 117, \textbf{MCS:} [(0, 1), 'inside'], [(0, 2), 'same-color'], [(0, 3), 'same-color'], [(1, 2), 'same-shape'], [(1, 3), 'same-shape']\\
\textbf{Count:} 37, \textbf{MCS:} [(0, 1), 'inside'], [(0, 2), 'same-color'], [(0, 3), 'same-color'], [(1, 2), 'same-shape']

\textbf{Task 11:}\\
\textbf{Count:} 196, \textbf{MCS:} [(0, 1), 'inside'], [(0, 2), 'same-color'], [(0, 3), 'same-color'], [(1, 2), 'same-shape'], [(1, 3), 'same-color'], [(2, 3), 'same-color']\\
\textbf{Count:} 117, \textbf{MCS:} [(0, 1), 'inside'], [(0, 2), 'same-color'], [(0, 3), 'same-color'], [(1, 2), 'same-shape'], [(1, 3), 'same-color']\\
\textbf{Count:} 82, \textbf{MCS:} [(0, 1), 'inside'], [(0, 2), 'same-color'], [(0, 3), 'same-color'], [(1, 2), 'same-shape']

\textbf{Task 12:}\\
\textbf{Count:} 15, \textbf{MCS:} [(0, 1), 'same-color'], [(0, 2), 'same-color'], [(1, 2), 'same-color'], [(1, 3), 'same-color'], [(2, 3), 'same-color']\\
\textbf{Count:} 9, \textbf{MCS:} [(0, 1), 'same-color'], [(0, 2), 'same-color'], [(0, 3), 'same-color'], [(1, 2), 'same-color'], [(1, 3), 'same-color'], [(2, 3), 'same-color']\\
\textbf{Count:} 2, \textbf{MCS:} [(0, 1), 'same-color'], [(0, 2), 'same-color'], [(0, 3), 'same-color'], [(1, 2), 'same-color']

\subsection{Task Maximum Common Subgraph Retrieval, with unseen tasks}
\label{appendix:task_mcs_preds_unseen}
The following show the Maximum Common Subgraph Retrievals for \proj trained on 2-3 core object, 1 distractor dataset with the contrastive + IB objective. The evaluation dataset is different, on the 2-4 core object, 0 distractor dataset. As a result, the relational structure of Tasks 6 to 12 are not seen by the model during training. 

\textbf{Task 0:}\\
\textbf{Count:} 420, \textbf{MCS:} [(0, 1), 'inside']

\textbf{Task 1:}\\
\textbf{Count:} 270, \textbf{MCS:} [(0, 1), 'inside'], [(0, 2), 'same-color'], [(1, 2), 'same-color']\\
\textbf{Count:} 150, \textbf{MCS:} [(0, 1), 'inside'], [(0, 2), 'same-color']

\textbf{Task 2:}\\
\textbf{Count:} 360, \textbf{MCS:} [(0, 1), 'inside'], [(0, 2), 'same-color'], [(1, 2), 'same-shape']\\
\textbf{Count:} 60, \textbf{MCS:} [(0, 1), 'inside'], [(1, 2), 'same-shape']

\textbf{Task 3:}\\
\textbf{Count:} 420, \textbf{MCS:} [(0, 1), 'same-color']

\textbf{Task 4:}\\
\textbf{Count:} 255, \textbf{MCS:} [(0, 1), 'same-color'], [(0, 2), 'same-color'], [(1, 2), 'same-color']\\
\textbf{Count:} 165, \textbf{MCS:} [(0, 1), 'same-color'], [(0, 2), 'same-color']

\textbf{Task 5:}\\
\textbf{Count:} 214, \textbf{MCS:} [(0, 1), 'same-color'], [(0, 2), 'same-shape'], [(1, 2), 'same-color']\\
\textbf{Count:} 206, \textbf{MCS:} [(0, 1), 'same-color'], [(0, 2), 'same-shape']

\textbf{Task 6:}\\
\textbf{Count:} 148, \textbf{MCS:} [(0, 1), 'same-color'], [(0, 2), 'same-color'], [(0, 3), 'same-color'], [(1, 2), 'same-color'], [(1, 3), 'same-color'], [(2, 3), 'same-shape']\\
\textbf{Count:} 146, \textbf{MCS:} [(0, 1), 'same-color'], [(0, 2), 'same-color'], [(0, 3), 'same-color'], [(1, 2), 'same-color'], [(1, 3), 'same-color']\\
\textbf{Count:} 79, \textbf{MCS:} [(0, 1), 'same-color'], [(0, 2), 'same-shape'], [(1, 2), 'same-color'], [(1, 3), 'same-color']

\textbf{Task 7:}\\
\textbf{Count:} 220, \textbf{MCS:} [(0, 1), 'inside'], [(0, 2), 'same-color'], [(0, 3), 'same-color'], [(1, 2), 'same-color'], [(2, 3), 'same-color']\\
\textbf{Count:} 116, \textbf{MCS:} [(0, 1), 'inside'], [(0, 2), 'same-color'], [(0, 3), 'same-color'], [(1, 2), 'same-color'], [(1, 3), 'same-color'], [(2, 3), 'same-color']\\
\textbf{Count:} 67, \textbf{MCS:} [(0, 1), 'inside'], [(0, 2), 'same-color'], [(0, 3), 'same-color'], [(1, 2), 'same-color']

\textbf{Task 8:}\\
\textbf{Count:} 135, \textbf{MCS:} [(0, 1), 'same-color'], [(0, 3), 'same-shape'], [(1, 2), 'same-color'], [(1, 3), 'same-color']\\
\textbf{Count:} 120, \textbf{MCS:} [(0, 1), 'same-color'], [(0, 2), 'same-shape'], [(0, 3), 'same-shape'], [(1, 2), 'same-color'], [(1, 3), 'same-color']\\
\textbf{Count:} 107, \textbf{MCS:} [(0, 1), 'same-color'], [(0, 2), 'same-shape'], [(0, 3), 'same-shape'], [(1, 2), 'same-color'], [(1, 3), 'same-color'], [(2, 3), 'same-shape']

\textbf{Task 9:}\\
\textbf{Count:} 137, \textbf{MCS:} [(0, 1), 'inside'], [(0, 2), 'same-color'], [(0, 3), 'same-color'], [(1, 2), 'same-color'], [(2, 3), 'same-shape']\\
\textbf{Count:} 128, \textbf{MCS:} [(0, 1), 'inside'], [(0, 2), 'same-color'], [(0, 3), 'same-color'], [(1, 2), 'same-color'], [(1, 3), 'same-color'], [(2, 3), 'same-shape']\\
\textbf{Count:} 108, \textbf{MCS:} [(0, 1), 'inside'], [(0, 2), 'same-color'], [(0, 3), 'same-color'], [(2, 3), 'same-shape']

\textbf{Task 10:}\\
\textbf{Count:} 258, \textbf{MCS:} [(0, 1), 'inside'], [(0, 2), 'same-color'], [(0, 3), 'same-color'], [(1, 2), 'same-shape'], [(1, 3), 'same-shape'], [(2, 3), 'same-shape']\\
\textbf{Count:} 112, \textbf{MCS:} [(0, 1), 'inside'], [(0, 2), 'same-color'], [(1, 2), 'same-shape'], [(1, 3), 'same-shape'], [(2, 3), 'same-shape']\\
\textbf{Count:} 46, \textbf{MCS:} [(0, 1), 'inside'], [(0, 2), 'same-color'], [(0, 3), 'same-color'], [(1, 2), 'same-shape']

\textbf{Task 11:}\\
\textbf{Count:} 201, \textbf{MCS:} [(0, 1), 'inside'], [(0, 2), 'same-color'], [(0, 3), 'same-color'], [(1, 2), 'same-shape'], [(1, 3), 'same-color'], [(2, 3), 'same-color']\\
\textbf{Count:} 98, \textbf{MCS:} [(0, 1), 'inside'], [(0, 2), 'same-color'], [(0, 3), 'same-color'], [(1, 2), 'same-shape'], [(1, 3), 'same-color']\\
\textbf{Count:} 93, \textbf{MCS:} [(0, 1), 'inside'], [(0, 2), 'same-color'], [(0, 3), 'same-color'], [(1, 2), 'same-shape']

\textbf{Task 12:}\\
\textbf{Count:} 16, \textbf{MCS:} [(0, 1), 'same-color'], [(0, 2), 'same-color'], [(1, 2), 'same-color'], [(1, 3), 'same-color'], [(2, 3), 'same-color']\\
\textbf{Count:} 9, \textbf{MCS:} [(0, 1), 'same-color'], [(0, 2), 'same-color'], [(0, 3), 'same-color'], [(1, 2), 'same-color'], [(1, 3), 'same-color'], [(2, 3), 'same-color']\\
\textbf{Count:} 2, \textbf{MCS:} [(0, 1), 'same-color'], [(0, 2), 'same-color'], [(0, 3), 'same-color'], [(2, 3), 'same-color']

\newpage
\subsection{Additional BabyARC examples}
\label{appendix:babyarc_task_more}
\centerline{\includegraphics[trim={0 1200 0 0},clip,width=\columnwidth]{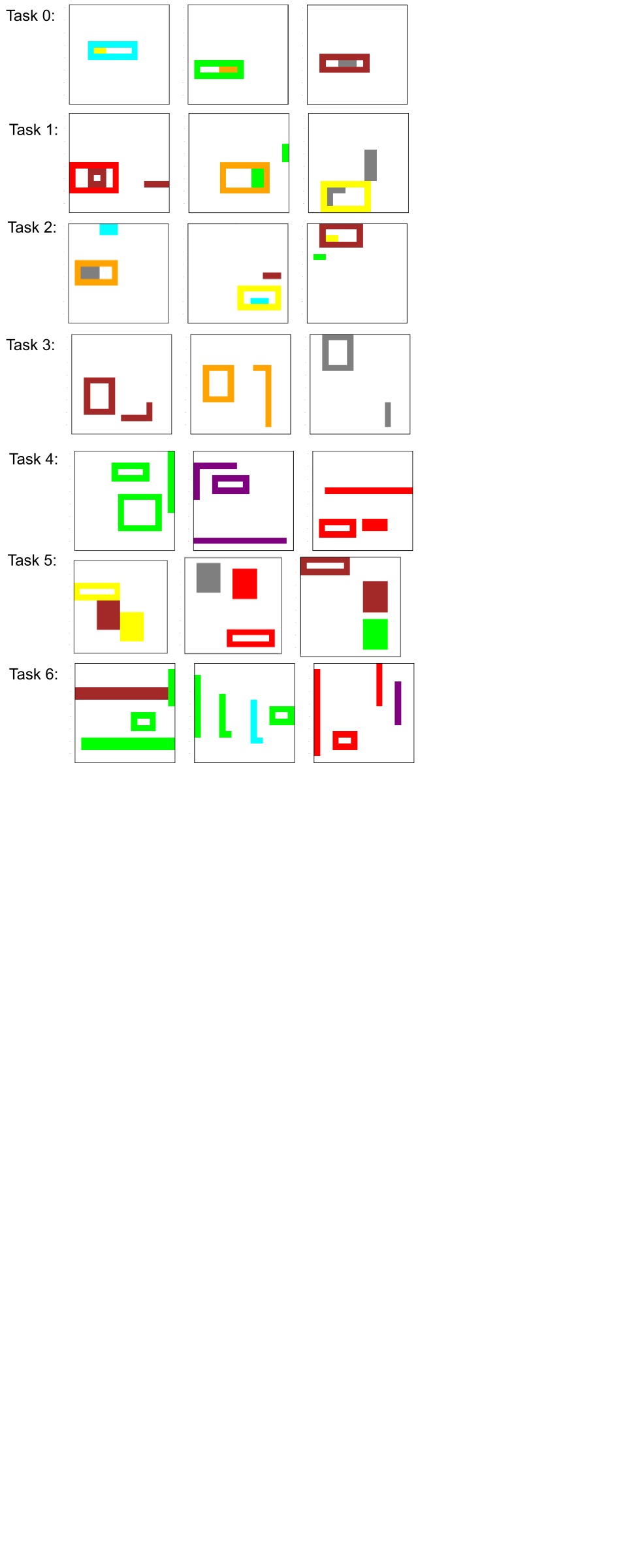}}

\newpage
\subsection{Additional ARC example tasks}
\label{appendix:arc_task_more}
\begin{figure}[h]
\begin{center}
\centerline{\includegraphics[width=0.5\columnwidth]{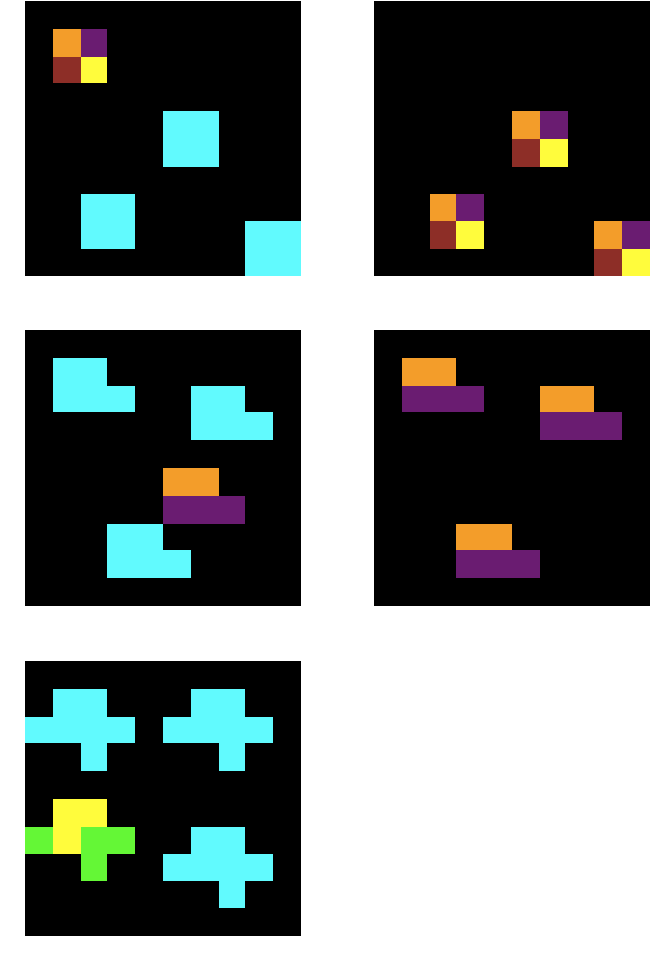}}
\caption{Example of an ARC task. The first column represents the input, and the second column represents the output, and the first two rows are independent training examples. The last row is the test example for this task, where only the input image is given. In this task, the objective is to color all the cyan objects with the same pattern as the non-cyan object, and then remove the original non-cyan object. This procedure explains obtaining the output from the input. All the training examples share a common relational graph structure, between the inputs and between the outputs. In the inputs, all the objects hold the relation ``same-shape``, and three of the objects hold the relation ``same-color``. In the outputs, all the three objects hold the relation ``same-shape`` and ``same-color``. The ability to learn and identify relations in unsupervised manner is essential, as this task requires both understanding of ``same-shape`` and ``same-color`` relations.}

\label{figure:arc_task}
\end{center}
\vskip -0.3in
\end{figure}

\begin{figure}[h]
\begin{center}
\centerline{\includegraphics[width=0.5\columnwidth]{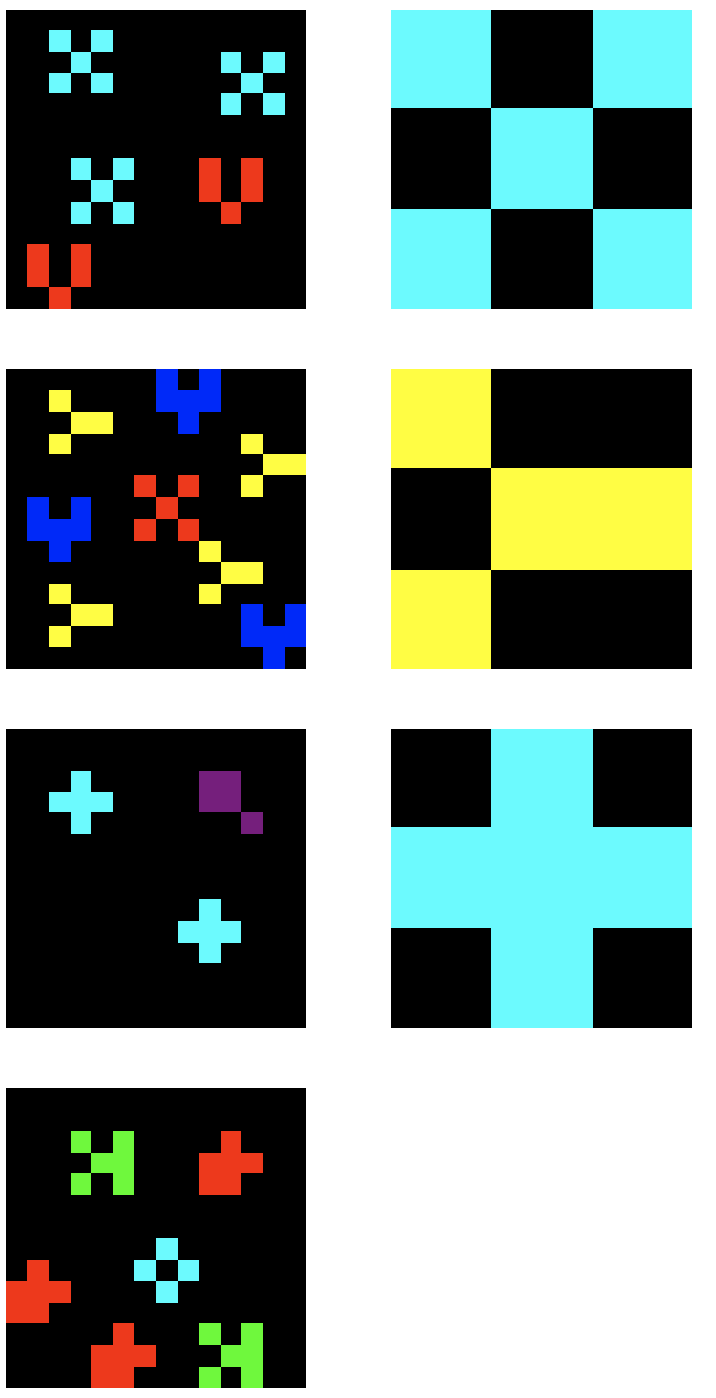}}
\caption{Another example of an ARC task. See caption in Fig. \ref{figure:arc_task} for general task description. In this task, the objective is to count the unique distinct objects by shape and color, and present the object with the largest count. In the inputs, the relational graph is the following: various groups of objects share the relation of ``same-color`` and ``same-shape`` with each other.}
\label{figure:arc_79}
\end{center}
\vskip -0.3in
\end{figure}

\begin{figure}[h]
\begin{center}
\centerline{\includegraphics[width=0.5\columnwidth]{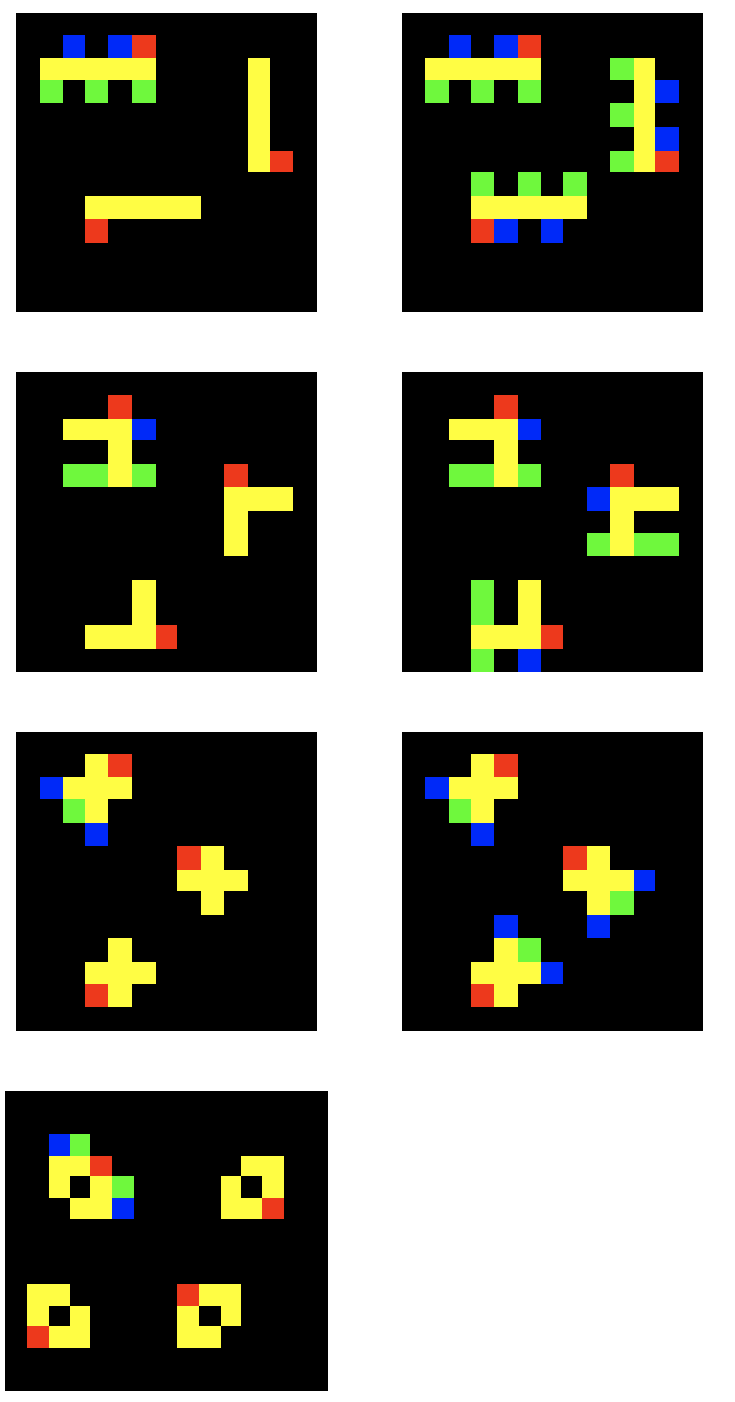}}
\caption{Another example of an ARC task. See caption in Fig. \ref{figure:arc_task} for general task description. In this task, the objective is identify the object with blue and green colors, and copy the object to the other shapes by aligning the red and yellow pixels. In the inputs, the relational graph is the following: all but one of the objects hold the relation ``same-shape`` and ``same-color``. In the outputs, all of the objects hold the relation ``same-shape`` and ``same-color``.}
\label{figure:arc_76}
\end{center}
\vskip -0.3in
\end{figure}

\end{document}